\def\year{2020}\relax
\documentclass[letterpaper]{article} 
\usepackage{aaai20}  
\usepackage{times}  
\usepackage{helvet} 
\usepackage{courier}  
\usepackage[hyphens]{url}  
\usepackage{graphicx} 
\usepackage{amsmath}
\usepackage{amssymb}
\usepackage{comment}
\usepackage{subfigure}
\usepackage{graphicx}  
\title{Recognizing Video Events with Varying Rhythms}
\author{Yikang Li\textsuperscript{\dag}, Tianshu Yu\textsuperscript{\dag}, Baoxin Li\\ Arizona State University\\ 
\{yikang.li,tianshuy,baoxin.li\}@asu.edu\\
\textsuperscript{\dag} indicates equal contribution
}
 
\begin{document}

\maketitle

\begin{abstract}
   Recognizing Video events in long, complex videos with multiple sub-activities has received persistent attention recently. This task is more challenging than traditional action recognition with short, relatively homogeneous video clips. In this paper, we investigate the problem of recognizing long and complex events with varying action rhythms, which has not been considered in the literature but is a practical challenge. Our work is inspired in part by how humans identify events with varying rhythms: quickly catching frames contributing most to a specific event. We propose a two-stage \emph{end-to-end} framework, in which the first stage selects the most significant frames while the second stage recognizes the event using the selected frames. Our model needs only \emph{event-level labels} in the training stage, and thus is more practical when the sub-activity labels are missing or difficult to obtain. The results of extensive experiments show that our model can achieve significant improvement in event recognition from long videos while maintaining high accuracy even if the test videos suffer from severe rhythm changes. This demonstrates the potential of our method for real-world video-based applications, where test and training videos can differ drastically in rhythms of sub-activities.
\end{abstract}

\section{Introduction}

In recent years, video-based analysis has brought about enormous and important challenges to computer vision, among which action recognition is a highlighted topic with practical importance. Many methods of different types (e.g. \cite{two-stream,trajectory-pooled,LCRN,TSN,Hierarchical1,Hieracrchical2}) have been proposed, reporting significant recognition results on several classic video recognition datasets such as UCF101 \cite{UCF101}, KTH \cite{KTH} and HMDB51 \cite{HMDB}, which contain simple \emph{actions} with small intra-class variations within a short time period. On the other hand, a more challenging task is to conduct event-level recognition on more complex datasets (e.g. VIRAT \cite{VIRAT} and Breakfast \cite{kuehne2014language}), where videos of much longer duration typically contain complex and/or multiple sub-activities.  To address this, several event recognition algorithms were proposed \cite{tran2008event,jiang2013high,PAMI_VIRAT,xu2015discriminative}, taking into account either longer-term dependency or the activity variation to some extent. However, an essential problem in long-video event recognition, which can greatly hinder the recognition effort, has never been considered: \emph{varying action/activity rhythms}. In this paper, we seek to design an \emph{end-to-end} deep framework to explicitly handle this issue.


Varying rhythm of actions in real videos may arise from at least two sources. First, the rhythm of sub-activities in an event can differ in nature. The period of sub-activities can be influenced by many internal and external factors. Consider, for example, the visual event ``getting into a car'' from the VIRAT dataset. One can either open the door and get into the car immediately, or open the door then hesitate for a while before entering the car. Though these two sets of actions are both categorized with the same label, their temporal rhythms on different activities differs significantly. In this case, an algorithm may fail to establish the effective temporal dependency when trained with one rhythm but tested with another. Second, the rhythm issue may occur due to non-uniform or different sampling rates between the training and testing stages. In practice, frames might be lost at arbitrary temporal positions (e.g., due to transmission loss), or a test video may be heavily re-sampled in time for real-time processing. In such cases, there may be dramatic sampling discrepancy among sub-activities in different video clips. In either of these cases, no existing algorithm can explicitly handle the varying-rhythm issue.


To address this, we propose an end-to-end deep framework, which is inspired in part by how humans identify events with changing rhythms. Intuitively, a human may fast forward unimportant segments of a video and slow down at critical parts to help abstracting essential event information. This importance evaluation procedure can be viewed as weight prediction or keyframe selection. Though there exist independent works on keyframe selection \cite{VS_LSTM,gong2014diverse,lee2012discovering} and event recognition \cite{tran2008event,jiang2013high,PAMI_VIRAT}, how to incorporate both to handle the varying-rhythm issue remains obscure. Our framework is, to the best of our knowledge, the first end-to-end system for this purpose. In our framework, we devise two lines of algorithms equipped with Recurrent Neural Networks (RNN) and Reinforcement Learning (RL) sub-routines. Our algorithm is trained with normal rhythm, but still can achieve stable performance with different rhythm in testing (simulated by re-sampling with varying frame rates). We also note that our model can handle very long videos (over 1,600 frames) where most existing approaches would fail.  In particular, the proposed SRNN+ algorithm achieves the best performance in long and complex video setting.

\begin{figure*}
\centering
    \includegraphics[width=0.9\textwidth]{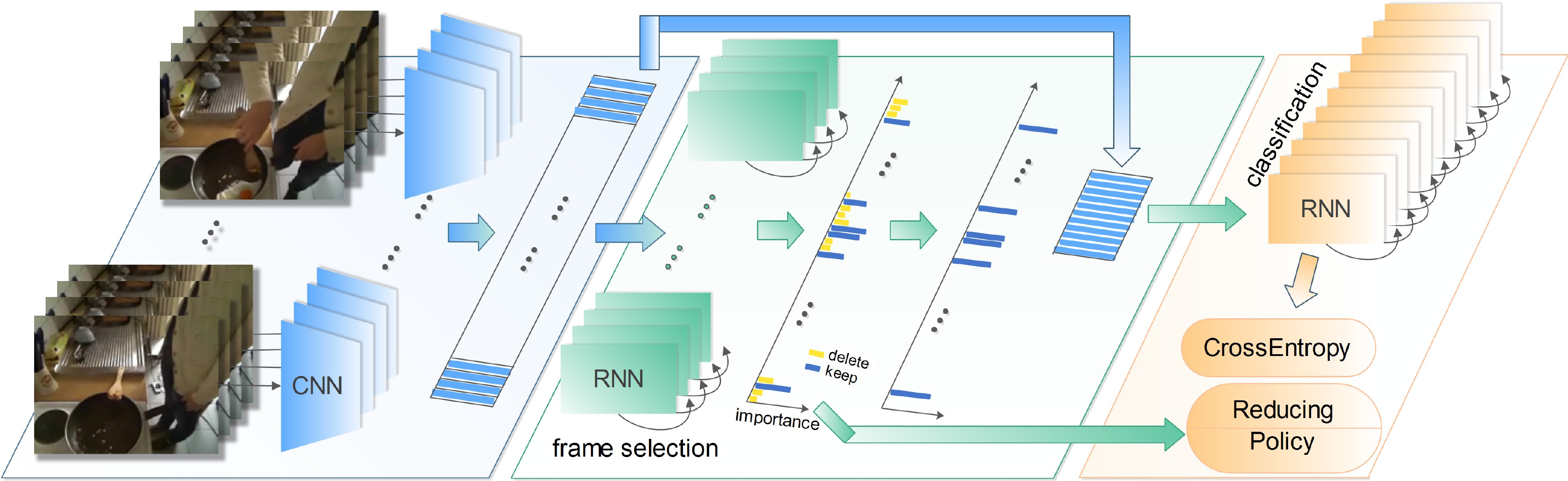}   
    %
    ~
    \caption{The illustration of our framework. The middle box and the right box correspond to frame selection and recognition modules, respectively. For the RNN-based algorithm, we employ the Reducing Regularization, while the Policy Function is adopted for the RL-based algorithm. For the classification RNN, we employ \textbf{GRU} in all our algorithms.}
    \label{fig:model}
\end{figure*}

\section{Related Work}

A large variety of neural-network-based algorithms has been proposed for different video tasks in recent years. In general, video action or event recognition algorithms based on neural networks can be roughly grouped into two categories: 1) Utilizing Convolutional Neural Networks (e.g. 3D CNNs or 2D CNNs) to obtain spatial-temporal features \cite{C3D,trajectory-pooled,two-stream,TSN}, where the temporal information is also encoded with convolutional operations; 2) Leveraging the Recurrent Neural Networks (e.g. GRUs or LSTMs) to obtain the temporal and order information from the extracted sequential features \cite{LCRN,action_atten,Hierarchical1,Hieracrchical2,li2018independently}. The second category of algorithms is considered to be more flexible in our case, since a singleton CNN structure requires fixed length for video segments to operate (e.g. in C3D \cite{C3D} the default lengths for 3D CNN is 16 frames), while RNN is designed to handle arbitrary sequence length.

Currently the performance of state-of-the-art action recognition models on many simple video datasets, such as KTH \cite{KTH} UCF101 \cite{UCF101}, can reach over $90\%$ accuracy. In the meantime, more complex video datasets \cite{VIRAT,kuehne2014language,ActivityNet} have also been built, aiming at capturing more sophisticated daily-life situations to help algorithmic development under more realistic challenges. In some research, the sampling rate is tuned as a singleton fixed parameter for better performance \cite{TSN,C3D,LCRN}. However, there is no discussion on the influence of the sampling rate gap between training samples and testing samples, which can be common in the real world. 

To process more complex videos, hierarchical or multiple RNNs \cite{Hierarchical1,Hieracrchical2,action_atten,li2018independently} were proposed to capture the higher-level structure of the sequential data, which is intended to reduce the sensitivity of lower levels of RNN. Some specific structures of RNNs were also proposed to handle longer sequence (e.g. IndRNN \cite{li2018independently}, QRNN \cite{bradbury2016quasi} and IRNN \cite{le2015simple}). However, the hierarchical structure would fail the task if the rhythm of testing sequences is beyond the higher-level step. As in this case, varying sampling rate or rhythm will violate the temporal dependencies the model has learned. We can also infer that the CNN based video action recognition models also suffer from the video rhythm problem. In \cite{TSN,C3D,3DCNN,two-stream}, the input length of the training and testing stages is fixed to an identical size. If the video rhythm changes, the kernels in CNN is not capable of capturing varying neighboring information. 

To our best knowledge, there is no previous work explicitly addressing the the general issue of varying action rhythms (due to either video sampling rate changes or natural variations in human actions), which can be common in real-world applications. 
Some analogous counterparts either predict frame importance independently, or perform recognition without selection. Video summarization, which can be viewed as keyframe or subshot selection procedure, is either supervised \cite{VS_LSTM,gong2014diverse} or unsupervised \cite{VS_gan,VS_RL,VS_alstm}. However, our framework evaluates the frame importance in a weakly-supervised fashion, as only event labels are given. Besides, the selected frames in our model do not necessarily reflect the importance of human perception. Instead, the selected frames are most contributing to the recognition task. Another line of methods based on deep reinforcement learning learns to select informative frames which is also similar to our work \cite{Progressive_RL,yeung2016end}. However, such algorithms require a reliable model to produce the actions. Therefore, the precondition of reinforcement learning is different from conventional action recognition algorithms. Some works \cite{blockdrop,VS_RL} also proposed to fine-tune or re-train the neural network along with the policy network. But once the neural network is re-trained, the whole model is getting more sensitive to the inputs, thus the video rhythm problem will likely happen. We also note that some action recognition algorithms perform random frame selection (e.g. TSN \cite{TSN}), but one cannot tell how much the selected frames contributes to the recognition.

\section{Methodology}

We employ a CNN+RNN-based structure to build our model, where CNN is to extract visual feature per frame and RNNs are in a hierarchical structure to select the frames and conduct event recognition. A schematic diagram of our model can be found in Figure \ref{fig:model}. The main idea of our frame selection procedure is to design a specific objective to reduce the required number of input frames, hence to automatically abandon the redundant sequential information. The dashed boxes in Figure \ref{fig:model} indicate the frame-selection procedure for both algorithms to be developed under the framework. As there is no supervised information telling which frames are more informative, our model employs a weakly-supervised mechanism to learn a frame-selection strategy given only the event-level labels. The CNN part of our algorithm is a VGG16 network \cite{VGG19} that is pre-trained on the ImageNet dataset. It produces a $4096$-dimensional feature for each frame and will \textbf{NOT} be updated during the whole training stage. We detail the algorithms with RNN and RL in the following.

\subsection{RNN-based Algorithm}
In this method, we utilize two layers of IndRNN \cite{li2018independently} to select the informative frames \cite{li2018independently} taking into account that we need to face relatively long video sequences, while IndRNN proved effective to handle gradient vanishing and exploding problem quite well against very long sequential data. Assuming that an input video sequence consists of frame data $\{\mathbf{x}_i\}$, where $i\in\{1, ... N\}$ presents the index of each frame, the updating rule of the hidden state of IndRNN can be written as:
\begin{equation}
    \mathbf{h}_t=\sigma\left(\mathbf{Wx}_t+\mathbf{u}\odot\mathbf{h}_{t-1}+\mathbf{b}\right)
    \label{equa:eq1}
\end{equation}
where $\odot$ is the Hardmard's product and $\sigma(\cdot)$ is the activation function. The classification RNN is a standard GRU \cite{cho2014learning}. After feeding the frame-selection module with the features, the output of this module is the sequence of decisions for each frame $\{y_i\|p_i\}$, where $y_i\in\{0,1\}$ and $p_i\in [0,1]$ represent the decision (being selected or not) and corresponding possibility (importance score) respectively. Namely $y_i = 1 \iff p_i \geq 0.5$ and  $y_i = 0 \iff p_i <0.5$ indicate ``keep'' and ``delete'', respectively. Following this instruction, the input video frames with ``keep'' tags are reformed into a much short video by our algorithms. Then the newly-formed video sequence is treated as the input for the following recognition/classification module. 

\textbf{Skip IndRNN} The aforementioned IndRNN can alleviate the gradient vanishing problem by independently aggregating the spatial patterns over time \cite{li2018independently}. However, the high dimension of the input ($4096$-dimensional data in our experiments) also increases the burden of the first IndRNN layer. In \cite{li2018independently}, the dimension of the input in all experiments is less than 50 (far less than 4096). We found that if we stack more layers of IndRNN (e.g. 6 layers), the output value of IndRNN will also increase by orders of magnitude (i.e. $>1e5$). To alleviate this, we proposed an improved IndRNN structure by skipping state updates to shorten the computational graph of IndRNN (\textbf{Skip IndRNN}). This idea is inspired by \cite{skiprnn} which implements skip operation on conventional RNN. Unlike \cite{skiprnn} whose \textbf{UPDATE} and \textbf{COPY} gate is computed by a matrix multiplication, our Skip IndRNN structure obtains the value of \textbf{UPDATE} and \textbf{COPY} gate by using Hardmard's product, which is the same as IndRNN. There are two advantages of utilizing Hardmard's product in computing the gate value. First, it keeps the independence of each neuron in IndRNN. Second, the gradient of the Skip IndRNN depends on the value of the weight instead of the weight matrix product. The mathematical representation of Skip IndRNN can be described as follows:
\begin{equation}
    \mathbf{u}_t = f_{binaries}({\Tilde{\mathbf{u}}}_t)
\end{equation}
\begin{equation}
    \mathbf{h}_t = \mathbf{u}_t \odot \mathbf{h}_{t} + (\mathbf{1} - \mathbf{u}_t) \odot \mathbf{h}_{t-1}
\end{equation}
\begin{equation}
    \Delta\mathbf{\Tilde{u}}_t = \sigma (\mathbf{w}_p \odot \mathbf{h}_t + \mathbf{b}_p)
\end{equation}
\begin{equation}
    \mathbf{\Tilde{u}}_{t+1} = \mathbf{u}_t\odot \Delta\Tilde{\mathbf{u}}_t+(\mathbf{1}-\mathbf{u}_t)\odot(\Tilde{\mathbf{u}} + \min(\Delta\Tilde{\mathbf{u}}_t, \mathbf{1}-\Tilde{\mathbf{u}}))
\end{equation}
where $\odot$ is the Hardmard's product and $\sigma(\cdot)$ is the sigmoid activation function. $f_{binaries}$ is the step function: $f_{binarize} : [0,1] \rightarrow \{0,1\}$ which binarizes the input value. $\mathbf{w}_p$ is the weight vector which can be learned to obtain the incremental value $\Delta\Tilde{\mathbf{u}}_t$. $\mathbf{h}_t$ and $\mathbf{h}_{t-1}$ is obtained by Eq (\ref{equa:eq1}).

Unlike \cite{skiprnn}, which updates the current state or copies previous state entirely, our Skip IndRNN updates or copies neurons independently. Some of the current neurons will be updated while some of the previous neurons will be copied directly. The \textbf{UPDATE} and \textbf{COPY} gate of each neuron will obtain the temporal relationship among the frame features. The \textbf{UPDATE} means the relationship between frames can be contributed to the temporal structure while the \textbf{COPY} leads to a skip action between frames which can be replaced by the more important ones during the learning stage.

To train our RNN-based model, we define a regularization term to enforce our model to select the least required number of informative frames. We call this regularization term as ``Reducing Regularization'' ($\mathcal{L}_R$) . This regularization term is added to the cross-entropy loss for classification to train the entire model. We introduce this term with more details in the following. 

\textbf{Reducing Regularization} ($\mathcal{L}_R$) is designed to reduce the required number of frames by making the output possibilities $P_i$ less than $0.5$ as much as possible. It can be described as following
\begin{equation}
    \mathcal{L}_R = \left|\frac{1}{N}\sum_{i=1}^N p_i-m_R\right|
\end{equation}
where the $m_R$ is the parameter that controls how much the possibility of ``delete" option $p_i$ is stronger than ``keep" option $1-p_i$. As $m_R$ becomes smaller, less frames tend to be selected for final recognition. One may suppose that with a low $m_R$, it can occur that no frame or very few frames are selected. However, we find this regularization term works well in practice, and produces proper number of the selected frames. We will discuss this term more in the experiments.

Together with the cross-entropy loss $\mathcal{L}_C$ for recognition, the final loss of this model can be written as:
\begin{equation}
    \mathcal{L}=\mathcal{L}_C+\lambda\mathcal{L}_R
\end{equation}
where $\lambda$ is a controlling parameter.

\subsection{RL-based Algorithm}
Inspired by the work of \cite{blockdrop,VS_RL}, we also design a frame-selecting algorithm employing reinforcement learning. As shown in Figure \ref{fig:model}, the policy network with RNN structure generates the decisions of ``keep'' or ``delete'' for the input videos and the rewards are computed jointly according to the usage of frames and the prediction accuracy of the recognition module.

\textbf{Policy Networks} The policy function is defined the same as in \cite{blockdrop}, which is a Bernoulli distribution related to the decision made for each frame. While the policy in \cite{blockdrop} is for dropping proper blocks of neural networks, our aim is to select the most informative frames. The policy function can be described as follows:
\begin{equation}
    \mathbf{\pi}(Y\|X) = \prod_{i=1}^{N}{p_i}^{Y_i}(1-p_i)^{1-Y_i}
\end{equation}
where as defined in the above section, $p_i$ is the probability of ``keeping'' the $i$th frame and $Y_i \in \{0,1\}$ indicates the action of the $i$th frame according to the probability. Only frames with a ``keep'' action are utilized for final recognition, which is the same procedure as in the RNN-based method.

Following the main idea of our models, which is to find the least required number of informative frames, the reward function is also designed with regularization on the usage of frames. Moreover, the accuracy of event recognition should also be considered for the policy network. A penalty parameter is set for the reward function which will penalize the network if the prediction is not correct \cite{blockdrop}. Thus the reward function $R$ can be written as:
\begin{equation}
    \mathcal{R} = 
    \begin{cases}
        1 - \left(\dfrac{K}{N}\right)^2 & \text{if correct} \\
        -\gamma & \text{otherwise}
  \end{cases}
\end{equation}
where $K$ and $N$ are the numbers of selected frames and total frames, respectively. And the penalty parameter $\gamma$ is introduced to control the trade-off between frame usage and recognition performance. 

\textbf{Gradient and Loss} The entire network (i.e. policy network + event recognition network) is trained with the cross-entropy loss from recognition and the reward loss form policy. As defined in \cite{blockdrop}, the expected reward should be maximized to optimize the parameters of the policy network. Thus the gradient of the policy network is derived as:  
\begin{equation}
    \nabla_\mathbf{W}\mathbb{E}_{{\mathbf{\pi}(Y\|X)}} = \mathbb{E}\left[\mathcal{R}\nabla_\mathbf{W}\sum_{i=1}^{N}\log[p_i Y_i + (1 - p_i)(1 - Y_i)]\right]
\end{equation}
where $\mathbf{W}$ is denoted as the weights of the policy network.

Moreover, as addressed in \cite{blockdrop}, the policy gradient methods are extremely sensitive to the weights initialization and not efficient if the policy is randomly initialized. Therefore, the curriculum learning \cite{CL} is applied during the training stage which proves effective on removing less informative frames for the event recognition model. The final loss for the RL-based network is defined as:
\begin{equation}
    \mathcal{L} = \mathcal{L}_C - \mathbb{E}_{\mathbf{\pi}(Y\|X)}[\mathcal{R}]
\end{equation}

\subsection{Analysis}
One may wonder how such a simple-structured network performs frame selection and how the effective gradient can be back-propagated. During each forward computation, the frame-selection layer generates a \emph{sparse connection} between frame importance module and recognition module dynamically controlled by reducing parameter $m_R$. During backward computation, the gradients are only propagated along the link in this sparse connection. At the early stage of the training, the frame prediction module almost makes a random guess on the frame importance. If this guess really matches real keyframes that matter most for the final recognition, the recognition module tends to make a smaller effort to learn the recognition strategy. In this case, the back-propagated gradients w.r.t the selection part are also small. Conversely, if the selected frames are less useful or even useless for recognition, the gradient will become much larger and the frame selection module will adjust the selection procedure. For a selected frame at time $t$, the gradient on RNN is back-propagated to all the time steps before $t$. In this fashion, our model can learn keyframe selection and event recognition simultaneously.

\section{Experiment}
\subsection{Datasets and Experimental Setup}
We demonstrate the impact of video rhythm problem and evaluate the proposed methods on three action (or event) recognition datasets, UCF101 \cite{UCF101}, VIRAT 2.0 surveillance video dataset \cite{VIRAT}, and Breakfast dataset \cite{kuehne2014language}, which represent, respectively, typical short, middle-range and long video clips for event/action recognition. All the experiments were conducted on a computer equipped with a \emph{single} GTX Titan Xp GPU with 12GB memory.

\textbf{UCF101 dataset} \cite{UCF101} consists of 101 action types with 13320 video clips. All of the video clips are trimmed from YouTube videos and each video clip contains only one action. We use the split 1 group, which is provided by \cite{UCF101} to separate training and testing samples.

\textbf{VIRAT 2.0 surveillance video dataset} \cite{VIRAT} includes about 8 hours of high-resolution surveillance videos (i.e. 1080p or 720p) with 12 kinds of events from 11 different scenes. In our experiment, we only focus on 6 types of person-vehicle interaction events occurring in a parking lot scene. The video clips are cropped based on the ground-truth bounding box. The training and testing video samples are randomly selected by following the ratio of 7:3.

\textbf{Breakfast dataset} \cite{kuehne2014language} comprises of 10 breakfast preparation related events that were performed by 52 different individuals in 18 different kitchen scenes. Each event was recorded by several cameras (i.e., $n=3\sim 5$ for different individuals in different scenes) and the total number of video clips is 1989. The overall video duration is about 77 hours and the average length of each video is about 140 seconds. We see the event in Breakfast dataset is far more complicated compared with those from UCF101 and VIRAT datasets. We split the dataset into training and testing by following the s1 split in \cite{kuehne2014language}.   

\textbf{Implementation Details}. We follow \cite{LCRN} to build a model with the CNN+RNN structure by leveraging Gated Recurrent Units (GRU). All the video frame features are 4096-dimensional pre-calculated vectors extracted from the FC2 layer of VGG16 \cite{VGG19}. The output dimension of the GRU is 1024. Following the GRU, there are two fully-connected layers with their respective number of neurons being 100 and the number of classes (i.e., n = 101, 6, 10 for UCF101, VIRAT, and Breakfast respectively). We employ softmax function in the final layer. 

For our RNN-based algorithm, we stack two Independent Recurrent Neural Networks (IndRNN) followed by two fully-connected layers as the frame-selecting model to generate possibilities of decision (importance) for each frame. The size of the hidden layer of IndRNN in the frame-selecting model is 250, and the numbers of neurons of the two fully-connected layers are 50 and 1, respectively. We term it \textbf{RNN+}. For our Skip IndRNN algorithm, we add one more vector weights to compute the incremental value of the update gate. The dimension of the vector weights is 250, which is the same as the hidden size of IndRNN layer in the RNN-based algorithm. We denote this setting \textbf{SRNN+}.

For our reinforcement-learning-based algorithm, the structure of the policy network is the same as the RNN-based algorithm, which consists of one IndRNN layer with two fully-connected layers. The parameter setting is the same as in the RNN-based method. And the penalty parameter that controls the trade-off between efficiency and accuracy is set to 1, which is the same as in \cite{blockdrop}. We name this algorithm \textbf{RL+}.

We implemented three baseline algorithms, \textbf{LRCN} \cite{LCRN}, \textbf{C3D} \cite{C3D}, and \textbf{TSN} \cite{TSN}, in a simple version with only spatial (RGB) features (without optical-flow). We also compared plain \textbf{IndRNN} \cite{li2018independently} for the Breakfast dataset. For C3D algorithms, we trained it from scratch on all three datasets and preprocessed the frames by following the method in \cite{C3D}. For LRCN and TSN, the frame feature was extracted from a pre-trained VGG16 network, which was not fine-tuned in the training stage. Only the RNN part of LRCN and the segmental consensus part of TSN were trained during the training stage.

Adam optimizer was utilized with the learning rate $1e-5$. $m_R=0.25$ in the Reducing Regularization term and the trade-off controlling parameter $\lambda=4$ empirically. The parameter $\gamma$ in the reward function was set to $1$. The training samples were collected by subsampling every two and three frames for the UCF101 and VIRAT datasets, respectively. For the Breakfast dataset, the training samples are subsampled by every 5 frames. The first and last 10 frames are removed from the training samples since those frames are mostly redundant according to the annotation of Breakfast dataset.

\subsection{Evaluation and Performance}
We demonstrate the impact of video rhythm problem via four scenarios, one of which is when the testing video sequences have the \textbf{same} sampling rate as the training inputs. The other three scenarios are designed with different kinds of sampling rates. To make the problem more challenging, we first equally divide each testing video into three intervals and apply different sampling rates to each interval to form a new testing sequence. For the scenario one (S1), we subsample the first and the third intervals with every two and five frames respectively to make those two periods more sparse, while keeping the rhythm intact for the middle interval. The testing inputs of scenario two (S2) are similar to S1 except we subsample the first and third intervals every five and two frames, respectively (reverse of S1). For the last scenario (S3), we randomly sample a half length of the testing video. Since the randomness of the last scenario brings uncertainty, we test the well-trained model 5 times and report the average performance of this scenario. We also provide the average frame usage (in percentage) of the whole testing dataset to demonstrate the efficiency of the frame selection mechanism.

\subsubsection{Performance on UCF101}
The performance on the UCF101 dataset is shown in Table \ref{tab:ucf}. The accuracy with the original sampling strategy of LRCN is comparable to the result cited from \cite{LCRN}, which is $68.19\%$. This shows the performance of our implementation of the baseline method with GRU on UCF101 is reasonable and the influence of video rhythm problem is convincing. The results of C3D and TSN indicate that the lack of sequential information (optical-flow) of the entire video will result in low accuracy. As the selected snippets would not differ too much in the testing stage, the gap of accuracy on different scenarios is within $1.5\%$. Compared with all three baselines, our proposed methods can achieve comparable results while maintaining a high accuracy on different sampling strategies. 

\begin{table}[tb]
    \centering
    \caption{Performance on UCF101 dataset. (\%)}
    \begin{tabular}{|c|c|c|c|c|c|}
\hline
         & original & S1 & S2 & S3 & Usage \\ \hline
C3D &   45.1 & 43.9 &  43.0  &  40.1  & -    \\ 
TSN &   56.9 & 56.3 &  55.4  &  56.0  & -    \\ 
LRCN &   69.2 & 60.7 &  65.3  &  62.5  & -    \\ 
RNN+  &   69.2 & 67.3 &  \textbf{68.7}  &  66.1  & 25.3   \\ 
SRNN+ &   68.9 & \textbf{67.9} &  \textbf{68.7}  &  \textbf{67.1}  & \textbf{25.1}   \\ 
RL+  &   \textbf{69.6} & 62.3 &  66.4  &  63.4  & 47.6   \\ \hline
\end{tabular}
    \label{tab:ucf}
\end{table}

\subsubsection{Performance on VIRAT}
The performance on the VIRAT dataset is shown in Table \ref{tab:virat}. The results demonstrate that the RNN-based algorithms performs much better than the CNN-based ones on the VIRAT dataset. The reason of the RNN-based being more favorable is that most of the training samples in VIRAT contain a collection of redundant sub-activities. We note that the performance degradation on S1-S3 has not been affected as much as in the UCF101 dataset. We believe that is due to the complexity and diversity of the VIRAT dataset. Some of the sub-activities (though not explicitly defined) can be inclusive by a specific event, but not by any others. For instance, the sub-action  ``lifting the loads'' is in ``Loading'' events exclusively. Therefore, if the varying rhythm does not violate those particular sub-actions, the event can still be classified correctly by the model. Regardless of the particularity of such sub-actions, the proposed methods can still maintain the performance under different subsampled scenarios.

\begin{figure*}[ht]
    \centering
    \includegraphics[width=.80\textwidth]{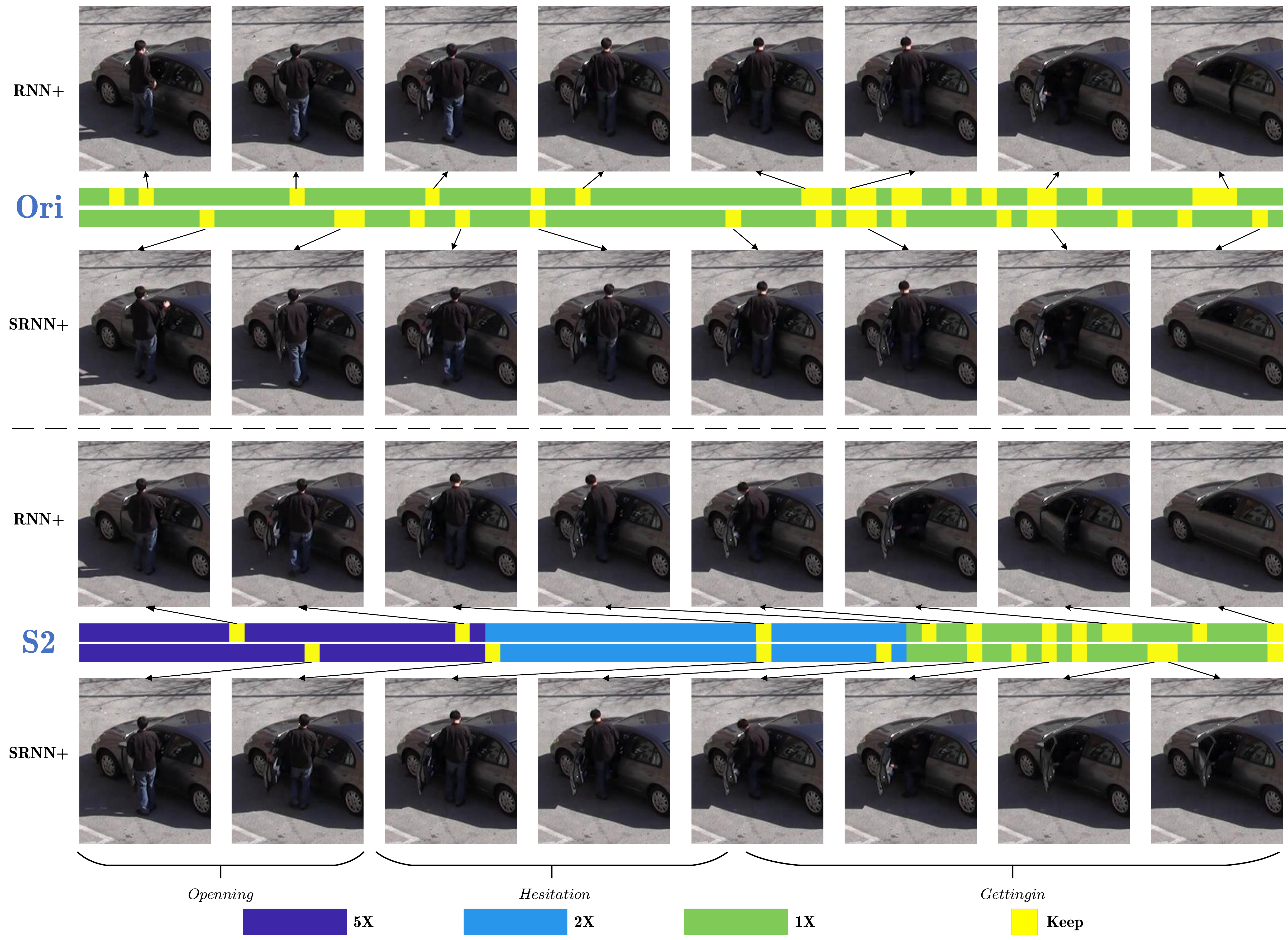}
    \caption{RNN+ and SRNN+ examples on the event ``getting into the vehicle'' with original (Ori) and S2 rhythm settings. Even under varying rhythm, our models are capable of skipping less informative frames and concentrate on informative ones.}
    \label{fig:virat_example}
\end{figure*}

An illustration of how the frame-selecting methods work is demonstrated in Figure \ref{fig:virat_example}. After adjusting the sampling rates, certain period of the event ``Getting into the Vehicle'' is missing, resulting in lack of sequential information. However, by applying the frame-selecting module to collectively abstract the most informative frames, our models can recognize the event from a few informative frames which are not necessarily adjacent in the original video. If at least some of those keyframes are not eliminated by subsampling, the frame-selecting model will effectively identify them and pass them to the recognition module. 

\noindent\textbf{Study of ${m}_R$}: Table \ref{tab:mr} further shows the performance of RNN+ with different ${m}_R$ values on the VIRAT dataset. We can find that the model performs best with ${m}_R = 0.3$ but the accuracy gap under different rhythm settings is about $4.4\%$. Therefore, we report ${m}_R = 0.25$ so as to reach a trade-off between accuracy of recognition and maintenance of performance. The behavior of SRNN+ with varying ${m}_R$ is similar to RNN+.

\begin{table}[t]
    \centering
    \caption{Performance on VIRAT dataset. (\%)}
    \begin{tabular}{|c|c|c|c|c|c|}
\hline
         & original & S1 & S2 & S3 & Usage \\ \hline
C3D &   42.9 & 40.2 &  37.7  &  41.1  & -    \\ 
TSN &   52.4 & 52.1 &  51.6  &  51.9  & -    \\ 
LRCN &   84.5 & 79.2 &  81.9  &  78.9  & -    \\ 
IndRNN &   82.6 & 80.3 &  81.2  &  81.0  & -    \\ 
RNN+  &   \textbf{84.9} & 82.3 &  82.0  &  81.5  & \textbf{23.9}   \\ 
SRNN+  &   84.7 & \textbf{83.9} &  \textbf{84.1}  &  \textbf{83.7}  & 25.7   \\ 
RL+  &   84.5 & 81.3 &  81.3  &  82.8  & 49.3   \\ \hline
\end{tabular}

    \label{tab:virat}
\end{table}

\begin{table}[t]
    \centering
    \caption{Performance on Breakfast dataset. (\%)}
    \begin{tabular}{|c|c|c|c|c|c|}
\hline
         & original & S1 & S2 & S3 & Usage \\ \hline
C3D &   10.8 & 10.8 &  10.8  &  10.8  & -    \\ 
TSN &   11.3 & 9.8 &  10.8  &  9.8  & -    \\ 
LRCN &   10.3 & 10.3 &  10.3  &  10.3  & -    \\ 
IndRNN  &   11.3 & 11.3 &  11.3  &  11.3  & -   \\ 
RNN+  &   29.4 & 19.6 &  20.6  &  17.2  & 25.4   \\ 
SRNN+  &   \textbf{32.9} & \textbf{24.6} &  \textbf{25.5}  &  \textbf{23.7}  & \textbf{24.1}   \\ 
RL+  &   30.8 & 16.7 &  17.6  &  16.2  & 49.2   \\ \hline
\end{tabular}
\renewcommand{\arraystretch}{1}

    \label{tab:bf}
\end{table}

\begin{figure*}[!ht]
    \centering
    \includegraphics[width=.80\textwidth]{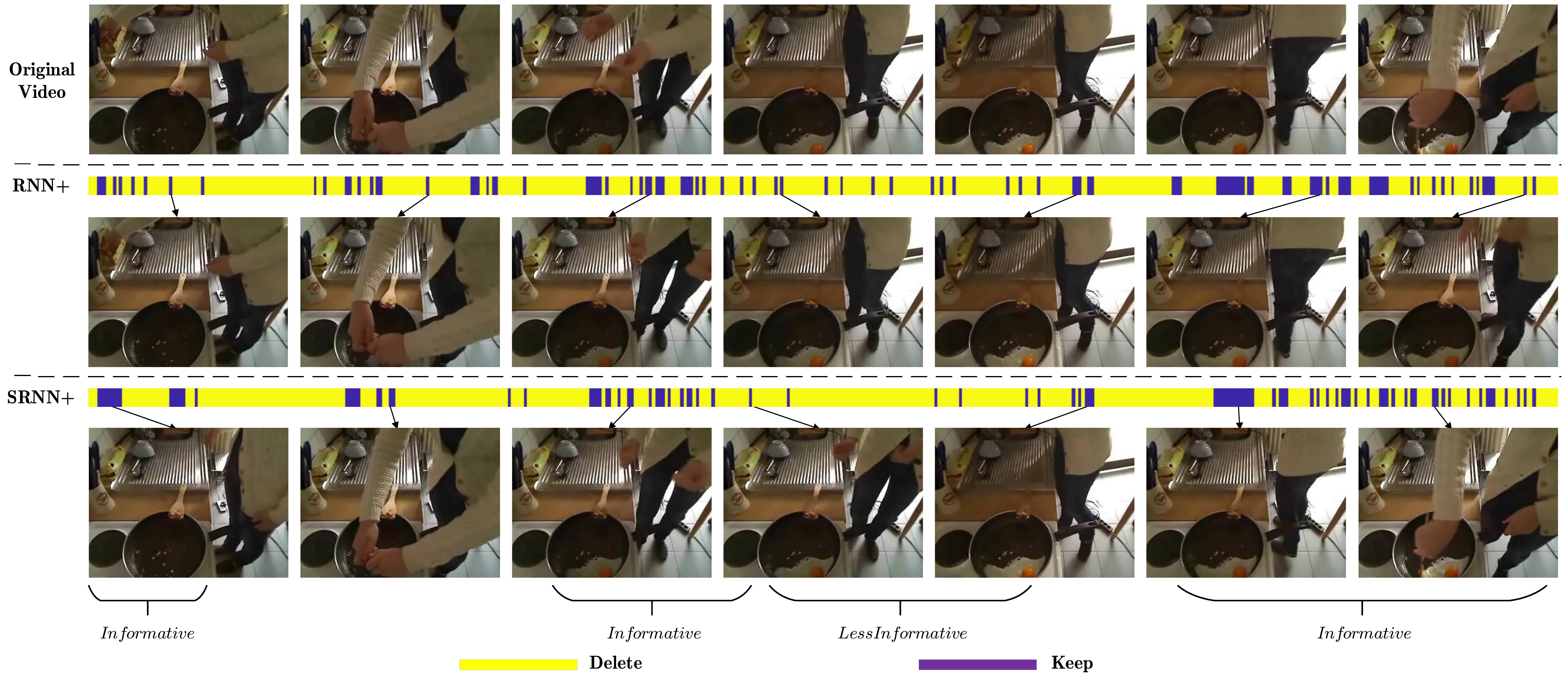}
    \caption{Frame selection and event recognition (frying egg) example with RNN+ and SRNN+. The first row of the frames corresponds to the original video sampled with equal interval (sampling 1 per 70 frames). The second and the third present the frame examples sampled from ``Keep'' tags by RNN+ and SRNN+, respectively. Zoom in for better view.}
    \label{fig:bf_example}
\end{figure*}

\subsection{Performance on the Breakfast dataset}
Table \ref{tab:bf} shows the performance on the Breakfast dataset. It can be seen that Breakfast is  more difficult than UCF101 and VIRAT. The average length of training videos in the Breakfast dataset is around 500 which is much longer than the average length in UCF101 ($\approx 70$) and VIRAT dataset ($\approx 170$). The maximal frame length can even go over 1600. Moreover, compared with the VIRAT dataset, the intra-class variations of sub-activities in the Breakfast dataset and the intrinsic rhythm changes over sub-activities are more severe, making it difficult or even ``impossible'' for the conventional methods (e.g., the selected baseline algorithms) to be well-trained on action recognition. We observe that all three baseline methods can hardly learn useful information from such sequences. However, by identifying informative frames, all the proposed algorithms can reach significant improvement (up to $32.9\%$) while the performance can be well maintained under varying sampling rate. In general, SRNN+ delivers the best performance on this dataset.

We visualize some examples in Figure \ref{fig:bf_example}. It can be noted that though RNN+ and SRNN+ differ in algorithmic details, both of them can learn strategies to utilize most informative frames. If we further see the two distributions of the selected frames, it can be concluded that these two algorithms make similar prediction on which frames are more informative and which are not, hence converge to a similar importance evaluation strategy. This fact cross-validates the effectiveness of the frame selection mechanism.

\begin{table}[tb]
    \centering
    \caption{RNN+ with varying ${m}_R$ on VIRAT dataset. (\%)}
    \begin{tabular}{|c|c|c|c|c|c|}
\hline
    ${m}_R$     & original & S1 & S2 & S3 & Usage \\ \hline
0.1 &   80.9 & 78.8 &  80.2  &  78.4  & 10.8    \\ 
0.2 &   80.2 & 81.1 &  80.7  &  79.9  & 21.1    \\ 
0.25 &  84.9 & 82.3 &  \textbf{82.0}  &  \textbf{81.5}  & 23.9    \\ 
0.3 &  \textbf{85.3} & \textbf{83.4} &  81.1  &  80.9  & 30.4   \\ 
0.4 &   82.4 & 81.1 &  81.5  &  79.6  & 39.9   \\ 
0.5 &   83.9 & 79.6 &  81.1  &  80.2  & 50.6   \\ \hline
\end{tabular}
\renewcommand{\arraystretch}{1}

    \label{tab:mr}
\end{table}

\section{Conclusion}
In this paper, we focus on investigating a new yet important problem, recognition with varying action rhythm, which intrinsically arises in real-world event-level videos and can greatly hinder the performance of existing methods. We demonstrated the influence of the rhythm problem by showing the performance of four baseline algorithms (i.e., C3D, TSN, LRCN and IndRNN) under four rhythm settings. All the baseline methods suffer from the video rhythms problem under long and complex video setting. We proposed two RNN-based frame-selecting models to help overcoming the video rhythm problem, as well as maintaining the accuracy. Especially, the proposed \textbf{Skip IndRNN} model can handle very long video sequences while maintaining the performance much better than any other baseline approaches.

\bibliographystyle{aaai}
\bibliography{ref}

\end{document}